\pdfoutput=1

\documentclass[11pt]{article}

\usepackage{acl}

\usepackage{times}
\usepackage{latexsym}

\usepackage[T1]{fontenc}

\usepackage[utf8]{inputenc}

\usepackage{microtype}

\usepackage{graphicx}
\usepackage{booktabs}
\usepackage{makecell}
\usepackage{multirow}
\usepackage{xurl}

\title{Overview of Robust and Multilingual Automatic Evaluation Metrics\\for Open-Domain Dialogue Systems at DSTC 11 Track 4}

\author{
    Mario Rodríguez-Cantelar\textsuperscript{\rm 1},
    Chen Zhang\textsuperscript{\rm 2},
    Chengguang Tang\textsuperscript{\rm 3},
    Ke Shi\textsuperscript{\rm 3},\\
    \textbf{
        Sarik Ghazarian\textsuperscript{\rm 4},
        João Sedoc\textsuperscript{\rm 5},
        Luis Fernando D'Haro\textsuperscript{\rm 1} and
        Alexander Rudnicky\textsuperscript{\rm 6}
    } \\
    \normalsize\textsuperscript{\rm 1}Speech Technology and Machine Learning Group - Universidad Politécnica de Madrid, Spain \\
    \normalsize\textsuperscript{\rm 2}National University of Singapore, Singapore \\
    \normalsize\textsuperscript{\rm 3}Tencent AI Lab, China \\
    \normalsize\textsuperscript{\rm 4}University of Southern California, USA \\
    \normalsize\textsuperscript{\rm 5}Department of Technology, Operations, and Statistics, New York University, USA \\
    \normalsize\textsuperscript{\rm 6}Carnegie Mellon University, USA \\
}

\begin{document}
\maketitle

\begin{abstract}
The advent and fast development of neural networks have revolutionized the research on dialogue systems and subsequently have triggered various challenges regarding their automatic evaluation. Automatic evaluation of open-domain dialogue systems as an open challenge has been the center of the attention of many researchers. Despite the consistent efforts to improve automatic metrics' correlations with human evaluation, there have been very few attempts to assess their robustness over multiple domains and dimensions. Also, their focus is mainly on the English language. All of these challenges prompt the development of automatic evaluation metrics that are reliable in various domains, dimensions, and languages. This track in the 11\textsuperscript{th} Dialogue System Technology Challenge (DSTC11) is part of the ongoing effort to promote robust and multilingual automatic evaluation metrics. This article describes the datasets and baselines provided to participants and discusses the submission and result details of the two proposed subtasks.
\end{abstract}

\section{Introduction}
\label{sec:introduction}

Recent advances in large-scale neural language models~\citep{devlin-etal-2019-bert, radford2019language, zhang-etal-2020-dialogpt} have led to significant attention in dialogue systems, especially in the open domain category. Significant research efforts are dedicated to boost the robustness of dialogue systems, that is, improving their capability to perform well across multiple domains, dimensions, and handling humans' diverse expressions of the same ideas (e.g., paraphrasing or back-translation). 

Automatic evaluation is an indispensable component for speeding up the development of robust dialogue systems. Common metrics are based on word overlap, such as BLEU~\citep{papineni-etal-2002-bleu} and ROUGE~\citep{lin-2004-rouge}, which mainly focus on matching syntactic information with a set of golden references. Unfortunately, such metrics correlate poorly with human judgments~\citep{liu-etal-2016-evaluate} as in open-domain dialogue, there can be limitless feasible responses w.r.t. a dialogue context. 

Alternatively, recently developed model-based metrics such as BERTscore~\citep{sun-etal-2022-bertscore}, BLEURT~\citep{sellam-etal-2020-bleurt}, FED~\citep{mehri-eskenazi-2020-unsupervised}, and MDD-Eval~\citep{zhang-etal-2022-mdd}, which take advantage of the strong semantic representation capability of pre-trained transformer language models, perform the evaluation at semantic and partially pragmatic levels. Some of them do not even need golden references as input. Regrettably, despite their improvement over the word-overlap metrics, these metrics are not perfect; that is, their correlation with human evaluation is still not strong. Moreover, most of them perform well only on a particular dimension (e.g., engagingness or coherence)~\citep{zhang-etal-2022-fined}, or specific to a single domain. In addition, their performance may be highly dependent on the datasets used for training and evaluation~\citep{yeh-etal-2021-comprehensive}.

Due to the lack of robust automatic evaluation metrics~\citep{mehri-eskenazi-2020-unsupervised}, researchers have to resort to the time-consuming and cost-intensive human evaluation process to analyze the performance of their model and benchmark their proposed methods against baselines.

Furthermore, to the best of our knowledge, none of the existing metrics have been thoroughly tested in a multilingual setting. Metric generalization across different languages is highly desirable, as it helps the transformation of state-of-the-art English-only dialogue systems into highly capable multilingual systems. Although multilingual pre-trained language models may exist and can be potentially used for training multilingual dialogue systems, human-annotations or high-quality dialogue datasets for languages other than English are very scarce or even nonexistent in the case of some low-resource languages. To address this problem, we take advantage of recent advances in neural machine translation and paraphrasing systems. Using existing high-quality services and models, it is possible to create new datasets for different languages and perform back-translation or paraphrasing to create additional data in the original language to improve and evaluate the robustness of existing metrics. To this end, we propose two subtasks in our track, and their details are listed as follows:

\subsection{Track Details}
\label{ssec:track_details}

This track consists of two tasks which are explained in more detail below.

Participants will develop effective open-ended and multilingual automatic dialogue evaluation metrics that perform similarly when evaluated in a new language. Participants will develop effective open-ended automatic dialogue evaluation metrics that perform robustly when evaluated over paraphrased/back-translated sentences in English. For both tasks, proposed metrics are expected to show the following two important properties, as indicated in~\citep{deriu2021survey}:

\begin{enumerate}
    \item Correlated to human judgments - the metrics should produce evaluation scores that well correlate to human judgments (scores) across multiple languages or alternative responses (i.e., back-translated or paraphrased).
    \item Explainable - the metrics should provide constructive and explicit feedback to the generative models in terms of the quality of their generated responses. For instance, if a generative model contradicts itself, the evaluation metrics should signal such behavior.
\end{enumerate}

Participants can propose their own metrics or optionally improve the deep AM-FM~\citep{zhang2021deep} baseline evaluation model provided by us. A leaderboard on the ChatEval platform\footnote{\url{https://chateval.org/dstc11}} was provided to check the performance of their different proposed models compared to those submitted by other researchers.

For each evaluation task, Spearman's correlation was used to compare the proposed evaluation metrics against human judgments. A final average score was calculated to rank the submitted metric models. Additional instructions to participants were provided through the Github repository\footnote{\url{https://github.com/Mario-RC/dstc11_track4_robust_multilingual_metrics}} and by email on the main DSTC distribution list.

\section{Task 1: Multilingual Automatic Metrics}
\label{sec:task1}

In this task, the goal for participants is to propose effective automatic dialogue evaluation metrics that exhibit the properties mentioned above (Section~\ref{ssec:track_details}) and perform well in a multilingual setup (English, Spanish, and Chinese). In concrete, participants were asked to propose a single multilingual model that could provide high correlations with human-annotations when evaluated in multilingual dialogues (development set in Section \ref{ssec:datasets-task1}) and perform well in the hidden multilingual test set. Participants were required to use pre-trained multilingual models and train them to predict multidimensional quality metrics using self-supervised techniques and, optionally, fine-tune their system over a subset of the development data.

Finally, participants evaluated their models on the development and test sets, expecting to show similar performance in terms of correlations with human-annotations across three languages: English, Spanish, and Chinese. Only development and test sets have human-annotations, and only the test sets were manually translated or paraphrased/back-translated to guarantee the correlations with the original human-annotations on the English data.

\subsection{Datasets}
\label{ssec:datasets-task1}

\textbf{Datasets summary}
Table~\ref{tab:all-data} shows the three clusters of datasets we used or created during the competition. The table shows information about the number of data used to train, develop, and test the proposed metrics. All these datasets clusters were available in English, Spanish, or Chinese and were back-translated into English. CHANEL and CDIAL include open-domain human-human conversations, while DSTC10 includes human-annotations on human-chatbot interactions. The type of  annotations or metadata and how each cluster was used (training/development/test) are indicated in the last three rows.

\begin{table*}[ht]
    \centering
    \resizebox{\textwidth}{!}{
    \begin{tabular}{cccc}
        \toprule
        \textbf{Dataset Name} & \textbf{CHANEL} & \textbf{DSTC10} & \textbf{CDIAL} \\
        \midrule
        \#datasets & 18 & 7 & 3 \\
        \midrule
        Language &
        \makecell[c]{English, Spanish/Chinese,\\
        and English back-translation} &
        \makecell[c]{English, Spanish/Chinese,\\
        and English back-translation} &
        \makecell[c]{English, Spanish/Chinese,\\
        and English back-translation} \\
        \midrule
        Dialogues Type & Human-Human Open-Domain & Human-Chatbot Open-Domain & Human-Human Open-Domain \\
        \midrule
        \#dialogues/utterances & + 390,000 / + 3,000,000 & + 18,000 / + 55,000	& + 3,470 / +130,000 \\
        \midrule
        Annotations & Sentiment analysis and Toxicity &
        \makecell[c]{Sentiment analysis and Toxicity \\
        Turn \slash dialogue level human scores} &
        Turn \slash dialogue level human scores \\
        \midrule
        Task 1 Set & Public: Train &
        \makecell[c]{Public: Dev, Test \\ Hidden: Automatic Translations} &
        Public: Train, Dev \\
        \midrule
        Task 2 Set & Public: Train	&
        \makecell[c]{Public: Dev, Test \\ Hidden: Manually back-translated\slash paraphrased} &
        — \\
        \bottomrule
    \end{tabular}
    }
    \caption{Summary of train\slash development\slash test datasets.}
    \label{tab:all-data}
\end{table*}

Table~\ref{tab:eval-data} (Appendix \ref{sec:appendixa}) provides a brief summary of all the statistics of the train, development, and test datasets. The datasets statistics including their number of utterances, avg. number of utterances in each conversation, avg. number of context/response words, type of annotations (turn or dialogue level), number of criteria, number of provided annotations, and type of dialogue systems used for generating responses are shown.

\textbf{Train}
As training set, we used the data released during the CHANEL@JSALT2020\footnote{\url{https://github.com/CHANEL-JSALT-2020/datasets}}\textsuperscript{,}\footnote{\url{https://www.clsp.jhu.edu/chaval-chat-dialogue-modeling-and-evaluation/}}~\citep{chanel-jsalt2020} workshop organized by Johns Hopkins University. This cluster consisted of a total of 18 well-known human-human dialogue datasets pre-processed and distributed in a standard format. The total number of dialogues was 393k (approximately 3M turns). An additional advantage of the data in this cluster is that they have been automatically translated back and forth using the same high-quality MS Azure translation service.\footnote{\url{https://azure.microsoft.com/en-us/products/cognitive-services/translator/}}


\textbf{Development}
As development set, the organizers provided data from two clusters of datasets: DSTC10 and CDIAL.

The first one was collected during DSTC10 Track 5~\citep{zhang-etal-2022-dstc10}, consisting of more than 35k turn-level human-annotations, which were automatically translated into Spanish and Chinese, and then back-translated into English using MS Azure services.

Second, we used datasets provided by THU-COAI\footnote{\url{https://github.com/thu-coai}} group (Conversational AI groups from Tsinghua University), naming this cluster of datasets CDIAL. It contains open-domain human-human dialogues. They are originally in Chinese and include 3,470 dialogues (approximately 130k turns). Furthermore, we provided Chinese to English translations through the SotA Tencent MT\footnote{\url{https://www.tencentcloud.com/products/tmt}} system.

Furthermore, Tencent AI manually annotated $\sim$ 3k random H-H turns ($\sim$1k dialogues) of CDIAL in Chinese (at turn-and dialogue-level).

It is important to note that the development data is intended to help participants verify the multilingualism and robustness capabilities of their models in terms of correlations with human-annotations.

\textbf{Test}
Furthermore, in order to check the generalization capabilities of the proposed metrics from the participant, the test data included new English, Chinese, and Spanish data of human-chatbot interactions (Appendix \ref{sec:appendixb}).

A new Human-Chatbot English dataset (HCEnglish) with $\sim$2k turns ($\sim$60 dialogues) with three different SotA chatbots (ChatGPT~\citep{radford2018improving}, GPT-3.5~\citep{brown2020language}, and BlenderBot 3~\citep{shuster2022blenderbot}~\citep{giorgi2023humancentered}). This dataset was manually annotated (turn-level and dialogue-level) using Amazon Mechanical Turk (AMT), then translated from English to Chinese and Spanish using MS Azure. 

In addition, a new Human-Chatbot Chinese dataset (HCChinese) consisting of $\sim$5k turns ($\sim$500 dialogues) was generated with three different SotA chatbots (Chinese DialoGPT, Microsoft's Xiaoice~\citep{zhou-etal-2020-design} and Baidu's Plato-XL~\citep{bao-etal-2022-plato}). This dataset was manually annotated (turn and dialogue level) by Tencent AI, and then translated from Chinese to English using the Tencent MT system.

Third, hidden data from the DSTC10 data was used for Spanish with a total of $\sim$1500 turns ($\sim$700 dialogues). Existing turn-level annotations were used, as well as Spanish translations and English back-translations created using MS Azure, which were subsequently manually reviewed.

Table \ref{tab:test-task1} shows the number of turns and dialogues for each test dataset for each language. The DSTC10 datasets did not have annotations at dialogue-level.

\begin{table*}[ht]
    \centering
    \resizebox{\textwidth}{!}{
    \begin{tabular}{ccccc|cccc|cccc}
        \toprule
        Language &
        \multicolumn{4}{c}{\textbf{EN}} &
        \multicolumn{4}{c}{\textbf{ZH}} &
        \multicolumn{3}{c}{\textbf{ES}} &
        \multirow{2}{*}{Global} \\
        Dataset &
        \textbf{HCEnglish} & \textbf{HCChinese} & \textbf{DSTC10} & \textbf{Total} &
        \textbf{HCEnglish} & \textbf{HCChinese} & \textbf{DSTC10} & \textbf{Total} &
        \textbf{HCEnglish} & \textbf{DSTC10} & \textbf{Total} & \\
        \midrule
        Turns     & 1,700 & 478 & 114 & 2,292 & 364 & 1,672 & 123 & 2,159 & 55 & 333 & 388 & 4,839 \\
        Dialogues & 59    & 40  & -   & 99    & 15  & 160   & -   & 175   & 3  & -   & 3   & 277   \\
        \bottomrule
    \end{tabular}
    }
    \caption{Summary statistics of the test dataset used for  \textbf{task 1} at turn and dialogue level, and separated by language.}
    \label{tab:test-task1}
\end{table*}

\textbf{Metadata}
Since the quality of translated sentences can play an important role in the estimation of metric scores, quality annotations between the original sentence and its respective translation were delivered for each turn of all datasets. Machine translation Quality Estimation (QE) metric scores were given to participants using the QE COMET\footnote{\url{https://github.com/Unbabel/COMET}}~\citep{rei-etal-2020-comet} system. In addition, for task 1, the cosine similarity between the original sentence and the translated sentence. Thanks to this information, participants could optionally discard dialogues or turns that potentially did not get a high translation quality estimation, therefore reducing potential noise but also allowing the creation of more robust metric systems.

In addition, toxicity and sentiment analysis metadata were provided for the original turns in both the CHANEL and DSTC10 datasets for filtering and dialogue curation purposes, as well as to avoid potential biases. These metadata allowed participants to have a better reference of the dataset quality, being of great help for them to decide whether or not to use these original turn and their translations in the training of their evaluation models and, optionally, fine-tune multilingual pre-trained models allowing better performance on the proposed dialogue-oriented tasks.

\textbf{Data Format}
All data given follow a unified data format to make the storage, handling, and retrieval easier for the participants. Detailed guidelines are available in the track repository.\footnote{\url{https://github.com/Mario-RC/dstc11_track4_robust_multilingual_metrics/blob/main/dstc11/track4-datasets-format.md}}

\textbf{Dimensions}
For HCEnglish, Amazon Mechanical Turk (AMT) was used to collect annotations for each of the dimensions evaluated in the test data. Our annotations restricted the users to location US, >97\% approval rate, >1000 HITs done, and a convenience pool of workers used for NLP evaluation tasks.\footnote{Without the convenience pool our annotator agreement was near random.} This pool included workers from the AMT filtering pipeline~\citep{zhang2022needle} and cloudresearch. The average compensation was $\sim$ 15\$/hr. We included text-based attention checks at the dialogue-level as well as an annotator agreement (both with an expert as well as between crowd workers [some from the DSTC10 dataset]) time-based filters on the turn-level.

For the HCChinese data, we leveraged the power of Tencent MT\footnote{\url{https://cloud.tencent.com/product/tmt}} to perform the English-to-Chinese translation of the corpus, followed by training a team of six professional Chinese annotators to annotate the dialogues. The entire annotation process spanned a month and incurred costs of approximately 6,194 US dollars, which is in line with the expenses associated with other evaluation datasets. The average cost of annotating each dialogue was 2.36 US dollars. 
Finally, the average correlation coefficient for Adequacy scored by six annotators is 0.79, and 0.67 for Fluency.

\subsection{Dimensions Evaluated}
\label{ssec:dimensions_evaluated-task1}

Since open-domain dialogue systems have multifacet nature, the evaluation can be accomplished from different perspectives. Since this is the case in both development and test data of task 1 (multilingual) and task 2 (robust), we include the following dimensions at turn-level and dialogue-level annotations~\citep{mehri2022report}:

\begin{itemize}
    \item[--] \underline{\textbf{Turn-level dimensions}}:\\
        \textbf{Appropriateness} - The response is appropriate given the preceding dialogue.\\
        \textbf{Content Richness} - The response is informative, with long sentences including multiple entities and conceptual or emotional words.\\
        \textbf{Grammatical Correctness} - Responses are free of grammatical and semantic errors.\\
        \textbf{Relevance} - Responses are on-topic with the immediate dialogue history.
    \item[--] \underline{\textbf{Dialogue-level dimensions}}:\\
        \textbf{Coherence} - Throughout the dialogue, is the system maintaining a good conversation flow.\\
        \textbf{Engageness/Likeability} - Throughout the dialogue, the system displays a likeable personality.\\
        \textbf{Informativeness} - Throughout the dialogue, the system provides unique and non-generic information.\\
        \textbf{Overall} - The overall quality of and satisfaction with the dialogue.
\end{itemize}

Furthermore, when choosing the test dimensions, the annotations available in the train and development data were taken into account to keep them balanced and homogeneous. 

The dimensions chosen at the turn-level show how much the responses are appropriate, informative including multiple entities and conceptual or emotional words, free of grammatical and semantic errors, and on-topic with the immediate dialogue history. The dimensions chosen at the dialogue-level show how much the system maintains a good conversation flow, engages well with the user, provides unique and non-generic information, and the overall quality of the system.

\begin{table}[ht]
    \centering
    \begin{tabular}{lcccc}
        \textbf{Sets} & \multicolumn{4}{c}{\textbf{Dimensions}} \\
        \midrule
        \underline{DSTC10} & & & & \\
        {\color[HTML]{6A9955} DSTC10-turn} &
        {\color[HTML]{6A9955} A} &
        {\color[HTML]{6A9955} CR} &
        {\color[HTML]{6A9955} GC} &
        {\color[HTML]{6A9955} R} \\
        {\color[HTML]{6A9955} ChatEval-turn} &
        {\color[HTML]{6A9955} A} & & & \\
        {\color[HTML]{6A9955} JSALT-turn} &
        {\color[HTML]{6A9955} A} & & & \\
        \midrule
        \underline{HCChinese} & & & & \\
        {\color[HTML]{4472C4} HCChinese-dial} &
        {\color[HTML]{4472C4} C} &
        {\color[HTML]{4472C4} EL} &
        {\color[HTML]{4472C4} I} &
        {\color[HTML]{4472C4} O} \\
        {\color[HTML]{4472C4} HCChinese-turn} & &
        {\color[HTML]{4472C4} CR} &
        {\color[HTML]{4472C4} GC} &
        {\color[HTML]{4472C4} R} \\
        \midrule
        \underline{HCEnglish} & & & & \\
        {\color[HTML]{FF0000} HCEnglish-dial} &
        {\color[HTML]{FF0000} C} &
        {\color[HTML]{FF0000} EL} &
        {\color[HTML]{FF0000} I} &
        {\color[HTML]{FF0000} O} \\
        {\color[HTML]{FF0000} HCEnglish-turn} &
        {\color[HTML]{FF0000} A} &
        {\color[HTML]{FF0000} CR} &
        {\color[HTML]{FF0000} GC} &
        {\color[HTML]{FF0000} R} \\
        \midrule
        \underline{Test data} & & & & \\
        {\color[HTML]{BF8F00} Test-dial} &
        {\color[HTML]{BF8F00} C} &
        {\color[HTML]{BF8F00} EL} &
        {\color[HTML]{BF8F00} I} &
        {\color[HTML]{BF8F00} O} \\
        {\color[HTML]{BF8F00} Test-turn} &
        {\color[HTML]{BF8F00} A} &
        {\color[HTML]{BF8F00} CR} &
        {\color[HTML]{BF8F00} GC} &
        {\color[HTML]{BF8F00} R} \\
    \end{tabular}
    \vskip 3mm
    \resizebox{\columnwidth}{!}{
    \begin{tabular}{r@{: }l r@{: }l}
    	$ C $  & Coherence               & $ EL $ & Engageness/Likeability \\
        $ I $  & Informativeness         & $ O $  & Overall \\
        $ A $  & Appropriateness         & $ CR $ & Content Richness \\
        $ GC $ & Grammatical Correctness & $ R $  & Relevance \\
	\end{tabular}
    }
    \caption{Summary of the dimensions (human-annotations) available for each dataset used in the test data, both at the turn and dialogue level.}
    \label{tab:test-dimensions}
\end{table}

Table \ref{tab:test-dimensions} summarizes the dimensions for each test data set. As can be seen, the DSTC10 set only has human turn-level annotations.

\subsection{Baseline}
\label{ssec:baseline-task1}

We provide a multilingual variant of deep AM-FM~\citep{zhang2021deep} (used previously during Track5 at DSTC10) as the baseline model. The formulation of both AM and FM remains unchanged except that we switch their original English-based pre-trained language models to multilingual models. For the adequacy metric (AM), we use XLM-R\footnote{\url{https://huggingface.co/sentence-transformers/xlm-r-100langs-bert-base-nli-stsb-mean-tokens}}~\citep{conneau-etal-2020-unsupervised} to extract sentence-level embeddings of both the response and the last sentence in the corresponding dialogue context. Then, the cosine similarity of the two embeddings is the AM score assigned to the corresponding response. For the fluency metric (FM), we adopt the multilingual GPT-2\footnote{\url{https://huggingface.co/ai-forever/mGPT}} as the backbone language model. The conditional probability of the response w.r.t. the context given by the multilingual GPT-2 model serves as the FM score of the response. The final AM-FM score is the arithmetic mean of both metric scores. All information related to the baseline model, such as code and data, can be found in this GitHub repository.\footnote{\url{https://github.com/karthik19967829/DSTC11-Benchmark}}




\subsection{Participants}
\label{ssec:participants-task1}

In Task 1, 4 teams participated, which provided a total of 16 submissions. Participants were asked to provide a brief description of the system for their proposals. The two system descriptions provided by the participants are shown below:

\textbf{Team 4}
Their approach utilizes two submetric groups, XLM-R and ChatGPT, for evaluating dialogue responses. The XLM-R group employs the XLM-Roberta-Large encoder model, consisting of NSP (Next Sentence Prediction), VSP (Valid Sentence Prediction), MLM (Masked Language Modeling), and ENG (Engagement) submetrics. The NSP submetric ensembles three models trained on English and multilingual data, while the VSP model combines different models. The ENG submetric uses an ensemble of encoder models trained on the ENDEX engagement dataset~\citep{xu-etal-2022-endex}. The MLM submetric utilizes the pre-trained XLM-R-large model with a Language Modeling head. The ChatGPT group prompts gpt-3.5-turbo to evaluate responses based on the dimensions of the DSTC11 test, with submetrics for dialogue and turn level. Weighted sums of the submetrics are calculated, with the weights learned from a subset of the dev dataset. For the test set, four variations were submitted, including weighted sums of XLM-R and ChatGPT, direct mapping of ChatGPT, and a weighted sum of all models.

In addition, Team 4 used the metadata provided. During their tests performed for task 1 they discovered that increasing the machine translated data affected the performance of the trained models. Therefore, they made use of the quality estimations computed with the COMET MTQE model to use only the best translated dialogues. 

For task 2, they trained their models using the least similar, and separately the most similar ones, based on cosine similarity and Levenshtein distance. They found that there was a good correlation between using the paraphrase score and their model performance, with lower scores bringing higher performance and vice versa. They deduced that the lower-scored responses were more diverse and therefore more informative for training.

\textbf{Team 7}
Their Parallel Corpus Alignment Framework enhances model evaluation on parallel corpora, focusing on Robust and Multilingual Automatic Evaluation Metrics for Open-Domain Dialogue systems. By utilizing xlm-roberta-large and bert-base as baseline models, they leverage representations from different languages, paraphrases, and translations to align parallel corpora in the semantic space. Through contrastive learning and multi-dataset distillation, they strengthen the model's scoring robustness and evaluation capability across various data domains.

\subsection{Results}
\label{ssec:results-task1}

Table~\ref{tab:task1-results-rank} shows the results on test data for task 1 at turn and dialogue level. To calculate the scores in the table, the following procedure was followed:
1. Data are separated in each language (English, Chinese, and Spanish);
2. Then, for each language separately, Spearman's correlation coefficients are calculated for each dimension independently;
3. Next, we calculate the mean of the correlations of the dimensions in each language (columns EN, ZH, and ES);
4. Finally, we calculate the final mean (Global column) of the language columns.


\begin{table}[h]
    \centering
    \resizebox{\linewidth}{!}{
    \begin{tabular}{cccccc}
        \multicolumn{6}{c}{\textbf{Turn-level}} \\
        \textbf{Team} & \textbf{EN} & \textbf{ZH} & \textbf{ES} & \textbf{Global} & \textbf{Rank} \\
        \midrule
        Baseline & 0.2940 & 0.0753 & 0.1826 & 0.1840             & 4 \\
        Team 2   & 0.1469 & 0.1054 & 0.0808 & 0.1110             & 5 \\
        Team 4   & 0.4818 & 0.3936 & 0.5890 & \textbf{0.4881}    & \textbf{1} \\
        Team 5   & 0.3702 & 0.0701 & 0.1983 & \textit{0.2129}    & \textit{3} \\
        Team 7   & 0.2214 & 0.3112 & 0.5644 & \underline{0.3657} & \underline{2} \\
        \midrule
        \midrule
        \multicolumn{6}{c}{\textbf{Dialogue-level}} \\
        \textbf{Team} & \textbf{EN} & \textbf{ZH} & \textbf{ES} & \textbf{Global} & \textbf{Rank} \\
        \midrule
        Baseline & 0.2414 & 0.4648 & 0.8080 & \underline{0.5047} & \underline{2} \\
        Team 4   & 0.5342 & 0.7133 & 0.8080 & \textbf{0.6852}    & \textbf{1} \\
        Team 5   & 0.1865 & 0.1356 & 0.6830 & \textit{0.3350}    & \textit{3} \\
    \end{tabular}
    }
    \caption{Spearman's correlations of the baseline and average correlations of each team's metrics on \textbf{turn-level} and \textbf{dialogue-level} test sets for \textbf{task 1}. The first position is shown in bold, the second in underline and the third in italics.}
    \label{tab:task1-results-rank}
\end{table}

To rank each team, the best submission was used according to the calculated global score. Teams 4, 7 and 5 were the best performers at turn-level. Regarding dialogue-level, team 4 was the best performer, followed by the baseline model and then by team 3. In particular, the performance of team 4 is outstanding in all languages.

This shows that team's 4 model is very effective not only at the global multilingual level, but also in each language separately, showing a very high performance in Spanish, followed by English and then Chinese. This highlights the need of a multilingual metric capable of performing in Chinese to match the results obtained in Spanish or English.

At dialogue-level, team 4 also demonstrated a very high correlation. Having a wide margin of advantage over team 5 and the base model. It should be noted that for Spanish at the dialogue-level, the amount of data was scarse, then producing non-statistical significant results and making difficult to analyze the reason for so high correlation results. 



\section{Task 2: Robust Evaluation Metrics}
\label{sec:task2}

In this task, the goal of the participants was to propose robust metrics for automatic evaluation of English dialogues that exhibit previously mentioned properties (subsection \ref{ssec:track_details}) while being robust when dealing with paraphrased/back-translated English sentences. Here, the expected performance for the proposed metrics was that they could be on par with the correlations with human-annotations obtained over the original sentences. As robustness criteria proposed, paraphrased/back-translated sentences should have the same semantic meaning as the original sentence but different wording. Task 2 was only evaluated for the English language.

Participants had the opportunity to evaluate their models with developmental data composed of paraphrased/back-translated sentences and their respective human annotations.

\subsection{Datasets}
\label{ssec:datasets-task2}

\textbf{Train, development, and test}
For task 2, the same task 1 datasets were used. However, to evaluate robustness, paraphrases and back-translated data were used. Thus, for task 2, the original datasets data was provided, in addition to the back-translations and paraphrases of the original sentences, but not the translations to other languages. Table \ref{tab:test-task2} shows the number of turns and dialogues for each test data set sent to the participants. The DSTC10 datasets did not have annotations at dialogue-level.

\begin{table}[ht]
    \centering
    \begin{tabular}{cccc}
        \toprule
        \textbf{Dataset} & \textbf{HCEnglish} & \textbf{DSTC10} & \textbf{Total} \\
        \midrule
        Turns     & 1,701 & 404 & 2,105 \\
        Dialogues & 59    & -   & 59    \\
        \bottomrule
    \end{tabular}
    \caption{Summary statistics of \textbf{task 2} test datasets at the turn and dialogue level.}
    \label{tab:test-task2}
\end{table}

For creating semantically similar sentences, we relied on two options: back-translations and a paraphraser model. For back-translations we used MS Azure MT services or Tencent MT system. Then for the paraphraser model, we used PARROT\footnote{\url{https://github.com/jsedoc/Parrot\_Paraphraser}}~\citep{prithivida2021parrot}. Multiple paraphrases were generated for all the original English sentences in each dataset.

For this task, paraphrases were preferable to back-translations. The reason is that current translation systems have a very high quality, so back-translations are often too similar to the original sentence, or even identical, not meeting in this case the robustness criterion proposed in task 2.

\textbf{Metadata}
For this specific task, participants received as metadata the Levenshtein\footnote{The Levenshtein distance is a numerical measure indicating the similarity between two strings. A higher Levenshtein distance signifies a greater difference between the two strings.} distance calculated for all paraphrases generated from the original sentences, in all datasets. For task 1, QE annotations were given using the same COMET model. In this case, they were calculated between the original sentence and its respective paraphrases separately, and between the original sentence and respective back-translation. 

Moreover, participants were given the Cosine Similarity calculation between the original sentence and its respective paraphrases, as well as between the original sentence and its back-translation. 
Finally, participants were notified of the provided metadata, as well as the toxicity and sentiment analysis annotations, for them to filter potentially biased or noised sentences.

\textbf{Dimensions}
Human-annotations for development and test data were the same as for task 1.

\subsection{Dimensions Evaluated}
\label{ssec:dimensions_evaluated-task2}

As the data for task 2 are the same as those in task 1, the nature of the data is common in both tasks. Therefore, the dimensions used to evaluate the models, both at the turn and dialogue level, are shared between the two tasks.

\subsection{Baseline}
\label{ssec:baseline-task2}

The same baseline was used for task 2, as for task 1\ref{ssec:baseline-task1}. In this case, paraphrases were used instead of multilingual sentences to evaluate robustness.

\subsection{Participants}
\label{ssec:participants-task2}
For this task, a total of 5 teams participated and sent a total of 21 submissions. Participants were asked to provide a brief description of their systems. Team 4 and 7 used the same models as for task 1, therefore they are descripted in Section \ref{ssec:participants-task1}. Below, we provide detailed description for teams 3 and 6.


\textbf{Team 3}
To address the variability of metrics in evaluating different dimensions and mitigate overfitting on scarce human-annotated data, they propose IDEL. This approach combines multiple metrics to achieve a higher correlation with human judgment across all dimensions. To avoid overfitting, they employed a list-wise learning-to-rank objective, leveraging the relative positions of examples rather than absolute coordinates. Furthermore, they utilized the LLaMa 65B dataset and the in-context-learning method for direct evaluation of examples, considering their context.


\textbf{Team 6}
Their approach focused on predicting turn-level qualities. They utilized pre-trained Large Language Models (LLMs) with manually designed prompts and two selected dialogues as few-shot examples to adapt the LLM output. Additionally, they built a feed-forward neural network (FNN) using frozen LLM representations as features to predict the desired metrics. Another submission employed the ChatGPT API with optimized prompts and dynamically obtained dialogue examples. Hyperparameters were selected based on manual annotations of 157 testing examples. However, for grammaticality metric scores, randomly generated scores were submitted due to uninformative constant scores predicted by the LLM.

\subsection{Results}
\label{ssec:results-task2}

Team results for turn and dialogue levels on the test data for task 2 are provided in Table~\ref{tab:task2-results-rank}. To calculate the scores and provide the ranking, the following procedure was followed:
1. Calculate the Spearman's correlation coefficients for each dimension independently;
2. Calculate the mean of the correlations of the dimensions;
3. Calculate the mean (Global column) of the language columns.


\begin{table}[h]
    \centering
    \resizebox{\columnwidth}{!}{
    \begin{tabular}{ccc|ccc}
        \multicolumn{3}{c}{\textbf{Turn-level}} & \multicolumn{3}{c}{\textbf{Dialogue-level}} \\
        \textbf{Team} & \textbf{Global} & \textbf{Rank} & \textbf{Team} & \textbf{Global} & \textbf{Rank} \\
        \midrule
        Baseline & 0.3387             & 4             & Baseline & \textbf{0.4800}    & \textbf{1} \\
        Team 1   & 0.1537             & 6             & Team 1   & 0.1111             & 4 \\
        Team 3   & 0.2697             & 5             & Team 3   & \textit{0.2196}    & \textit{3} \\
        Team 4   & \textbf{0.4890}    & \textbf{1}    & Team 4   & \underline{0.3031} & \underline{2} \\
        Team 6   & \underline{0.4190} & \underline{2} & & & \\
        Team 7   & \textit{0.3833}    & \textit{3}    & & & \\
    \end{tabular}
    }
    \caption{Spearman's correlations of the baseline and average correlations of each team's metrics on \textbf{turn-level} and \textbf{dialogue-level} test sets for \textbf{task 2}. The first position is shown in bold, the second in underline and the third in italics.}
    \label{tab:task2-results-rank}
\end{table}

The best presentation according to the overall score calculated was used to rank each team. Teams 4, 6 and 7 were the best performers at the turn-level. At dialogue-level, the baseline model provided the best performance, followed by team 4 and team 3.

Considering team 4 results, both in task 1 and task 2, it can be considered their model as the overall best in the competition, being good at multilingual level as well as in robustness. However, the performance of the baseline model at dialogue-level is far superior to that of team 4, showing there is still room for improvement.

\section{Conclusions and Future Work}
\label{sec:conclusions}

This paper presents a comprehensive overview of Track 4 on "Robust and Multilingual Automatic Evaluation Metrics for Open-Domain Dialogue Systems" organized as part of the 11\textsuperscript{th} Dialogue System Technology Challenge (DSTC11). The track was divided into two subtasks addressing an important problems in Dialogue Systems: the design of automatic evaluation metrics for multilingual dialogues and dialogue robustness when dealing with paraphrases or back-translations.

First task was divided at turn and dialogue level. At the turn-level, 4 teams actively participated and at the dialogue-level 2 teams participated. Having some of the teams participated at both levels. Some of the teams obtained interesting results that effectively contribute to the state-of-the-art of automatic evaluation models for multilingual dialogues. However, the results at the language level show a disparate performance among the different languages, giving room for improvement in the evaluation of other languages. The overall performance of the participants was satisfactory, with some teams outperforming the baseline model both in language and globally, as well as at the turn and dialogue levels. However, we can see that the automatic evaluation is still an open problem as correlation scores are still below 0.7 in the best of the cases.

The second task was also subdivided at turn and dialogue level. At the turn-level, 5 teams actively participated and at dialogue-level 3 teams, with some of the teams having participated at both levels. At the turn-level, several teams outperformed the baseline model. However, no team was able to outperform the baseline model at dialogue-level, showing that there is still room for improvement.

As future work, we plan to increase the number of databases, as well as to provide better baseline models. We also want to include a larger number of dimensions so that the evaluations performed are more complete, covering more different aspects of the dialogue. For task 1, it is planned to extend the number of available languages, to create multilingual models with a wider spectrum, thus widening the scope of the competition and attracting more participants who are fluent in other languages. For task 2 we want to propose higher quality paraphrases, such as those generated with models like GPT-4~\citep{openai2023gpt4}.

\section*{Acknowledgements}

This work is supported by project BEWORD (PID2021-126061OB-C43) funded by MCIN/AEI/10.13039/501100011033 and, as appropriate, by “ERDF A way of making Europe”, by the “European Union”, and by  the European Commission through Project ASTOUND (101071191 — HORIZON-EIC-2021-PATHFINDERCHALLENGES-01).
We gratefully acknowledge valuable efforts from Tencent AI Lab who supports Chinese translation and annotation of datasets by funding and infrastructure.
Thanks to THU-CoAI (Conversational AI groups from Tsinghua University) for providing their Chinese datasets as part of the challenge data.
Thanks to Unbabel for providing the COMET MTQE scores annotations as part of the challenge data. This contribution was supported by national funds through Fundação para a Ciência e a Tecnologia (FCT) with references PRT/BD/152198/2021 and UIDB/50021/2020, and by the P2020 program MAIA led by Unbabel (LISBOA-01-0247-FEDER-045909).
We also give thanks to MS Azure services (especially to Irving Kwong) for their sponsorship to continue processing new datasets for the research community.
This research project is supported by the NYU ChatEval Team led by João Sedoc.
This research project is supported in part by a grant from Amazon to Alexander Rudnicky, Carnegie Mellon University.
Thanks to Karthik Ganesan, Sarik Ghazarian, James Hagerty, Zhang Chen and Alex Rudnicky for developing the baseline model as part of the challenge tasks.

\bibliography{anthology,custom}
\bibliographystyle{acl_natbib}

\appendix

\section{Appendix: Datasets statistics}\label{sec:appendixa}

Table \ref{tab:eval-data} shows all the data sets that make up the training, development, and test sets. In detail, it shows the number of dialogues, turns per dialogue, average number of turns per dialogue, average number of words per turn, as well as the granularity of the annotations (at turn and/or dialogue level), the original language of the dataset and into which languages it is translated.

\begin{table*}[ht]
    \centering
    \resizebox{\textwidth}{!}{
        \begin{tabular}{l|ccccccc}
        \toprule
        \multirow{2}{*}{Name} &
        \multirow{2}{*}{\#Turns} &
        \multirow{2}{*}{\#Dialogues} &
        \multirow{2}{*}{\begin{tabular}[c]{@{}c@{}}Average\\Turn/Dial\end{tabular}} &
        \multirow{2}{*}{\begin{tabular}[c]{@{}c@{}}Average\\Words/Turn\end{tabular}} &
        \multirow{2}{*}{\begin{tabular}[c]{@{}c@{}}Annotation\\Granularity\end{tabular}} &
        \multirow{2}{*}{\begin{tabular}[c]{@{}c@{}}Original\\Language\end{tabular}} &
        \multirow{2}{*}{Translation} \\ \\
        \midrule
        
        \textbf{\underline{Train}} & & & & & & & \\
        DBDC~\citep{higashinaka-etal-2016-dialogue} & 8,509 & 415 & 20.5 & 7.31 & Turn & En & Zh/Es \\
        CMU\_DoG~\citep{zhou-etal-2018-dataset} & 95,305 & 4,221 & 22.58 & 17.93 & Turn & En & Zh/Es \\
        Cornell Movie-Dialogs~\citep{Danescu-Niculescu-Mizil+Lee:11a} & 304,713 & 83,097 & 3.67 & 13.72 & Turn & En & Zh/Es \\
        DailyDialog~\citep{li-etal-2017-dailydialog} & 102,960 & 13,116 & 7.85 & 13.96 & Turn & En & Zh/Es \\
        DECODE~\citep{nie-etal-2020-decode} & 296,105 & 35,426 & 8.36 & 15.05 & Turn & En & Zh/Es \\
        EmotionLines~\citep{hsu-etal-2018-emotionlines} & 14,503 & 1,000 & 14.50 & 10.53 & Turn & En & Zh/Es \\
        EmpathicDialogues~\citep{rashkin-etal-2019-towards} & 107,220 & 24,850 & 4.31 & 15.88 & Turn & En & Zh/Es \\
        Holl-E~\citep{moghe-etal-2018-towards} & 91,452 & 9,071 & 10.08 & 17.74 & Turn & En & Zh/Es \\
        MEENA~\citep{adiwardana2020humanlike} & 3,675 & 193 & 19.04 & 9.14 & Turn & En & Zh/Es \\
        MELD~\citep{poria-etal-2019-meld} & 23,197 & 1,592 & 14.57 & 10.98 & Turn & En & Zh/Es \\
        MetalWOz~\citep{lee2019multi} & 432,036 & 37,884 & 11.40 & 8.47 & Turn & En & Zh/Es \\
        Movie-DiC~\citep{banchs-2012-movie} & 512,582 & 65,215 & 7.86 & 13.82 & Turn & En & Zh/Es \\
        PersonaChat~\citep{zhang-etal-2018-personalizing} & 162,064 & 10,907 & 14.86 & 11.72 & Turn & En & Zh/Es \\
        SentimentLIAR~\citep{upadhayay2020sentimental} & 12,781 & 12,781 & 1.00 & 20.16 & Turn & En & Zh/Es \\
        Switchboard Coherence~\citep{cervone-riccardi-2020-dialogue} & 12,059 & 1,000 & 12.06 & 20.55 & Turn & En & Zh/Es \\
        Topical-Chat~\citep{Gopalakrishnan2019} & 235,281 & 10,784 & 21.82 & 23.23 & Turn & En & Zh/Es \\
        Wizard of Wikipedia~\citep{dinan2019wizard} & 201,999 & 22,311 & 9.05 & 18.83 & Turn & En & Zh/Es \\
        Wochat~\citep{haro2016intestinal} & 19,881 & 607 & 32.75 & 6.75 & Turn & En & Zh/Es \\
        \midrule
        Total & 2,636,322 & 334,470 & 236.26 & 255.77 & & & \\
        \midrule
        \textbf{\underline{Development}} & & & & & & & \\
        ConvAI2-GRADE~\citep{huang-etal-2020-grade} & 1,800 & 600 & 3.0 & 12.07 & Turn & En & Zh/Es \\
        DailyDialog-GRADE~\citep{huang-etal-2020-grade} & 900 & 300 & 3.0 & 12.60 & Turn & En & Zh/Es \\
        DailyDialog-GUPTA~\citep{gupta2019investigating} & 2,460 & 500 & 4.92 & 12.37 & Turn & En & Zh/Es \\
        DailyDialog-ZHAO~\citep{zhao-etal-2020-ddzhao} & 4,248 & 900 & 4.72 & 12.41 & Turn & En & Zh/Es \\
        DSTC7~\citep{galley2019grounded} & 34,650 & 9,990 & 3.47 & 15.39 & Turn & En & Zh/Es \\
        Empathetic-GRADE~\citep{huang-etal-2020-grade} & 900 & 300 & 3.0 & 16.65 & Turn & En & Zh/Es \\
        FED-Dial~\citep{mehri-eskenazi-2020-unsupervised}) & 1,715 & 125 & 13.72 & 11.1 & Dial & En & Zh/Es \\
        FED-Turn~\citep{mehri-eskenazi-2020-unsupervised}) & 3,888 & 375 & 10.37 & 10.78 & Turn & En & Zh/Es \\
        HUMOD~\citep{merdivan2020human} & 37,468 & 9,499 & 3.94 & 7.97 & Turn & En & Zh/Es \\
        Persona-SEE~\citep{see-etal-2019-makes} & 39,792 & 3,316 & 12.0 & 9.0 & Dial & En & Zh/Es \\
        PersonaChat-USR~\citep{mehri-eskenazi-2020-usr} & 2,790 & 300 & 9.3 & 12.08 & Turn & En & Zh/Es \\
        PersonaChat-ZHAO~\citep{zhao-etal-2020-ddzhao} & 4,614 & 900 & 5.13 & 12.06 & Turn & En & Zh/Es \\
        TOPICAL-USR~\citep{mehri-eskenazi-2020-usr} & 4,032 & 360 & 11.2 & 23.16 & Turn & En & Zh/Es \\
        ECM-Eval~\citep{zhou-etal-2018-emotion} & 3,004 & 1,502 & 2.0 & 13.13 & Turn & Zh & En \\
        KdConv-Eval~\citep{zhou-etal-2020-kdconv} & 3,499 & 354 & 9.88 & 21.11 & Turn & Zh & En \\
        LCCC-Eval~\citep{wang2020large} & 3,009 & 589 & 5.11 & 11.72 & Turn & Zh & En \\
        \midrule
        Total & 148,769 & 29,910 & 104.76 & 212.64 & & & \\
        \midrule
        \textbf{\underline{Test}} & & & & & & & \\
        BlenderBot3~\citep{giorgi2023humancentered,shuster2022blenderbot} & 679 & 21 & 32.33 & 16.96 & Turn/Dial & En & Zh/Es \\
        ChatGPT~\citep{giorgi2023humancentered,radford2018improving} & 462 & 21 & 22 & 91.07 & Turn/Dial & En & Zh/Es \\
        GPT-3.5~\citep{giorgi2023humancentered,brown2020language} & 560 & 17 & 32.94 & 23.73 & Turn/Dial & En & Zh/Es \\
        HCChinese & 2,017 & 187 & 10.79 & 8.08 & Turn/Dial & Zh & En \\
        ChatEval~\citep{sedoc-etal-2019-chateval} & 400 & 200 & 2 & 8.13 & Turn & En & Zh/Es \\
        DSTC10~\citep{zhang-etal-2022-dstc10} & 112 & 28 & 4 & 14 & Turn & En & Zh/Es \\
        JSALT~\citep{chanel-jsalt2020} & 46 & 13 & 3.54 & 17.26 & Turn & En & Zh/Es \\
        \midrule
        Total & 4,276 & 487 & 107.60 & 179.23 & & & \\
        \bottomrule
        \end{tabular}
    }
    \caption{Summary of the train, development, and test datasets. Some information comes from~\citet{yeh-etal-2021-comprehensive}.}
    \label{tab:eval-data}
\end{table*}

\section{Appendix: Existing Benchmark Datasets}
\label{sec:appendixb}

Descriptions of the datasets that constitute the DSTC10 benchmark can be found at~\citet{zhang-etal-2022-dstc10}. Details of the remaining evaluation datasets are as follows:

\bigskip
\noindent \textbf{ECM-Eval} - The test instances in ECM-Eval test set are sampled from the Emotional Short-Text Conversation (ESTC) dialogue corpus~\citep{zhou-etal-2018-emotional}, which is built on top of the Short-Text Conversation (STC) dataset~\citep{shang-etal-2015-neural}. ESTC is designed to build Chinese empathetic dialogue systems. The dialogues are crawled from Weibo and post-processing, such as the removal of trivial responses and filtering out potential advertisements, has been conducted by~\citet{shang-etal-2015-neural}. The dialogues are automatically annotated by pre-trained emotion classifiers along six different emotion categories, such as angry, happy, sad, etc. The dialogues in ESTC are much shorter. Most contain only a single post-response pair.

\bigskip
\noindent \textbf{LCCC-Eval} - Data in LCCC-Eval are sampled from the Large-scale Cleaned Chinese Conversation dialogue corpus (LCCC)~\citep{wang-etal-2020-large}. The LCCC corpus is designed for pretraining the Chinese dialogue model. The dialogues are mainly collected from Weibo, a Chinese microblogging website\footnote{\url{https://en.wikipedia.org/wiki/Sina_Weibo}} and other open-source Chinese dialogue corpora, such as the Douban Conversation~\citep{wu-etal-2017-sequential} and the E-Commerce Conversation Corpus~\citep{zhang-etal-2018-modeling}. All the dialogues belong to the general domain and a rigorous cleaning process, which is based on a series of heuristic rules and several classifiers, is conducted to filter out dialogues with noise, such as dirty words, special characters, facial expressions, ungrammatical sentences, etc. Both ESTC and LCCC are released by the THU-COAI group for research purposes at \url{https://www.luge.ai/\#/} 

\bigskip
\noindent \textbf{KdConv-Eval} - KdConv-Eval is constructed based on the KdConv corpus~\citep{zhou-etal-2020-kdconv}, a multi-domain Chinese dialogue dataset towards
multi-turn knowledge-driven conversation. The corpus links the subjects of multi-turn discussions to knowledge graphs. It encompasses conversations from three categories (movies, music, and travel). These conversations involve detailed exchanges about relevant subjects and seamlessly move between a variety of topics. We sampled 354 dialogues from the original corpus to form the KdConv-Eval test dataset.

\bigskip
\noindent \textbf{HCChinese} - Dialogues in HCChinese are collected by interacting with three state-of-the-art Chinese chatbots, Baidu Plato-XL~\citep{bao-etal-2022-plato},  Microsoft XiaoIce~\citep{zhou-etal-2020-design}, and a Chinese DialoGPT model that is trained in a similar manner to DialoGPT~\citep{zhang-etal-2020-dialogpt}. We chat with the chatbots on a diverse set of topics, such as entertainment, relationship, arts, travel, food, etc. The discussion of sensitive topics, such as politics and race, was avoided. A manual check is performed on each dialogue, and those containing inappropriate responses were filtered out. In the end, we collected 531 human-chatbot multi-turn conversations with 207 from Plato-XL, 224 from XiaoIce, and 100 dialogues from the Chinese DialoGPT.

\bigskip
\noindent \textbf{TBD-Q1-2023}
Three Bot Dialog Evaluation Corpus (TBD-Q1-2023 OR TBD; Quarter 1 of 2023) from~\citet{giorgi2023humancentered} consists of dialogues with three chatbots: ChatGPT~\citep{radford2018improving}, GPT-3~\citep{brown2020language}, and BlenderBot3~\citep{shuster2022blenderbot}. Student participants were told to have a long conversation with the chatbots on a range of topics of their choosing.
All conversations were collected between November 2022 and March 2023.
They collected 21 dialogues with an average of 14.6 turns per dialogue. 
Conversations for \textbf{BlenderBot3} were directly from the website~\url{https://blenderbot.ai/}. The \textbf{ChatGPT} conversations were taken directly from the website~\url{https://chat.openai.com/}. Finally, \textbf{GPT3} used the text-davinci-003 model with the prompt \textit{Hal is a chatbot that attempts to answer questions with useful responses:}. The GPT-3 model parameters were temperature of 0.5, max tokens of 289, top-p of 0.3, frequency penalty of 0.5, and presence penalty of 0. 

\end{document}